\def\year{2020}\relax
\documentclass[letterpaper]{article} %
\usepackage{aaai20}  %
\usepackage{times}  %
\usepackage{helvet} %
\usepackage{courier}  %
\usepackage[hyphens]{url}  %
\usepackage{graphicx} %
\usepackage{graphicx}  %
\title{AI Trust in business processes: The need for process-aware explanations}
\author{
Steve T.K. Jan\textsuperscript{\rm 1},
Vatche Ishakian\textsuperscript{\rm 2},
Vinod Muthusamy\textsuperscript{\rm 2} \\
\textsuperscript{\rm 1}Virginia Tech,
\textsuperscript{\rm 2}IBM Research AI
}

\begin{document}

\maketitle

\section{Introduction} 
\label{sec:intro}

Business processes underpin a large number of enterprise operations including loan origination, invoice management, and insurance claims processing~\cite{van2011process}. The business process management (BPM) industry is expected to approach \$16 billion by 2023~\cite{marketwatch}.
There is a great opportunity for infusing AI to reduce cost or provide better customer experience~\cite{rao2017sizing}, and the BPM literature is rich in machine learning solutions to gain insights on clusters of process traces~\cite{nguyen2016process,nguyen2019summarized}, predict outcomes~\cite{breuker2016comprehensible}, and recommend decisions~\cite{mannhardt2016decision}. Deep learning models including from the NLP domain have also been applied~\cite{tax2017predictive,evermann2017predicting}.

Unfortunately, very little of these innovations have been applied and adopted by enterprise companies~\cite{daugherty2018human},
and those adopted are limited to narrow domains such as customer services, enterprise risk and compliance~\cite{wilson2016companies}.

We assert that a large reason for the lack of adoption of AI models in BPM is that business users are risk-averse and do not implicitly trust AI models. There has been little attention paid to \emph{explaining model predictions to business users with process context}. These business users are typically experts in their fields but not data scientists, and explanations must be presented in their business domain vocabulary. We challenge the BPM community to build on the AI interpretability literature, and the AI Trust community to take advantage of business process artifacts.

\section{Example of process-aware explanations} 
\label{sec:example}

\begin{figure}[t] \centering
\includegraphics[width=\linewidth]{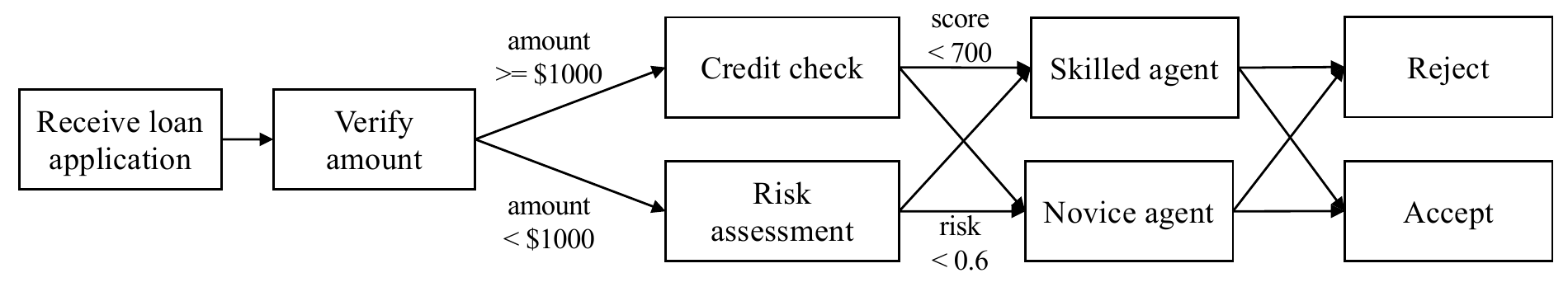}
\caption{Example loan application business process}
\label{fig:loanprocess}
\end{figure}

\begin{figure}[b]
\begin{minipage}[t]{0.48\linewidth}
    \centering
    \includegraphics[width=.8\textwidth]{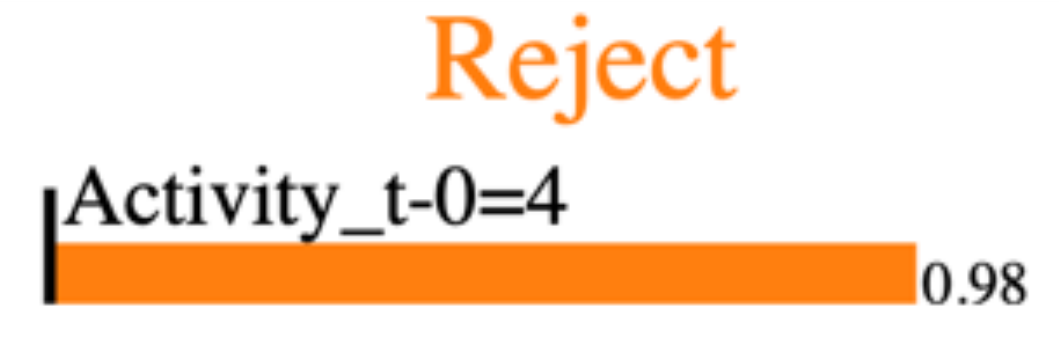}
    \caption{Vanilla LIME: The explanation is incomplete showing only that the loan is rejected because it is sent to a skilled agent.}
    \label{fig:lime-nocausal}
\end{minipage}
\hfill
\begin{minipage}[t]{0.48\linewidth} 
    \centering
    \includegraphics[width=.8\textwidth]{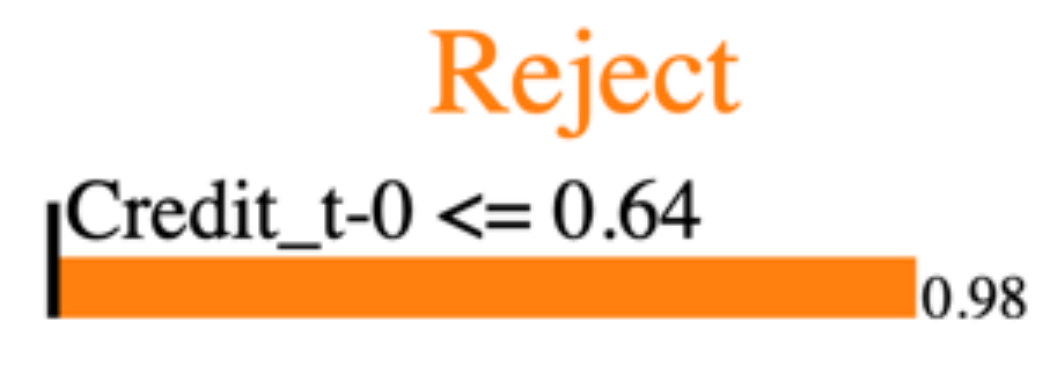}
    \caption{Process-aware LIME: The explanation now highlights the influence of the causal \textsc{credit score} feature.}
    \label{fig:lime-causal}
\end{minipage}        
\end{figure}

Consider the example loan application process in Fig.~\ref{fig:loanprocess}. 
Suppose we build a sequence model that takes as input the activities and features observed in the process and predicts the outcome, in this case whether a loan will be approved. Such models have been shown to achieve accuracies of up to 85\%~\cite{evermann2017predicting}. We can then use tools such as LIME~\cite{ribeiro2016should} to explain the prediction, and as we see in Fig.~\ref{fig:lime-nocausal}, LIME suggests that the presence of the skilled agent activity causes the application to be rejected.

A subject matter expert, however, would understand from the process description that large loan requests from borrowers with low credit scores are the ones most likely to end up routed to a skilled agent. This is an example of a causal relationship that can be inferred from the process description: the \textsc{loan amount} and/or \textsc{credit score} features are the cause of the feature associated with the skilled agent activity.

The explanations from LIME are based on sampling perturbations around the input features and measuring how the predictions change with the perturbations. 
It turns out in our example, that many of the perturbations do not conform to the process description and hence can never occur. Once we apply the causal relationship above to constrain the perturbation sampling, LIME offers the explanation in Fig.~\ref{fig:lime-causal} where the \textsc{credit score} is now prominent.

This simple example illustrates how directly applying interpretability techniques to process models results in incomplete or potentially misleading explanations. We also see how by being process-aware we can augment existing algorithms to improve the quality of explanations.
This is a nascent and fertile research area that we have only scratched the surface of.

\section{Challenges to trustworthy AI for BPM} 
\label{sec:challenges}

\begin{figure}[h] \centering
\includegraphics[width=\linewidth]{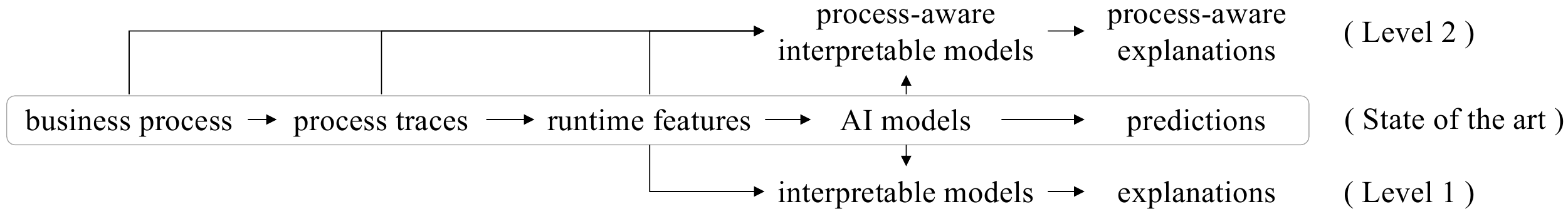}
\caption{Approaches for interpretability in BPM AI models}
\label{fig:roadmap}
\end{figure}

As stated earlier, virtually all AI models in the BPM literature train models with features from process traces, depicted as ``state of the art'' in Fig.~\ref{fig:roadmap}. %

There is an opportunity to apply known interpretability approaches in a BPM context, denoted as \textsc{Level 1} in Fig.~\ref{fig:roadmap}.
For example, a regression model to predict process completion time~\cite{polato2014} can be augmented by techniques to make regression models more interpretable~\cite{regressioninterpretability}.
The same can apply to deep learning models for process prediction tasks. Most of the current techniques, however, are based on sequence models including LSTMs and RNNs~\cite{tax2017predictive,evermann2017predicting}. While there has been a lot of research on explaining deep learning models based on CNNs for image classification tasks~\cite{gancvpr2018}, there is a  growing interest in similar problems for sequence models~\cite{nlpinterpretability}. LEMNA attempts to offer high fidelity explanations for deep learning models~\cite{wenboccs2018} but assumes a security context where adjacent features are dependent on each other, an assumption that does not always hold true in business processes.
Note that \textsc{Level 1} interpretability only uses the trained model (either as a black box or white box) and possibly features from the training data. Of course, applying known interpretability techniques to the new business process domain may not be straightforward and require additional innovations, but \textsc{Level 1} at least offers researchers a pathway to begin experimenting.

A more ambitious, and promising, approach is to bring \emph{process-awareness} to the problem, marked as \textsc{Level 2} in Fig.~\ref{fig:roadmap}. Here, interpretability models would take advantage of the knowledge of the business process definitions and full runtime process traces. Some of the information in these artifacts is typically lost when preparing the data for the predictive models. There is, unfortunately, a dearth of solutions that apply this approach. An example of \textsc{Level 2} was presented in the previous section where the black box interpretability model in LIME is augmented with knowledge of the causal relationships derived from the business process definition. An understanding of the feature causality graph avoids misleading or incomplete explanations.

Business process datasets\footnote{ \url{https://data.4tu.nl/repository/collection:event_logs_real}} are publicly available, but as part of this research area the community will need to develop metrics to measure the quality of the explanations. Standard techniques that measure how model accuracy degrades as features are removed from the dataset may make sense in other machine learning domains, such as visual explanations of image classification models~\cite{saliencymaps}, but may not be appropriate for business process.

We think process-aware explanations is an interesting research area with potential for high business impact.

{
\small
\bibliographystyle{aaai}
\bibliography{bibtex}

\begin{thebibliography}{}

\bibitem[\protect\citeauthoryear{Adebayo \bgroup et al\mbox.\egroup
  }{2018}]{saliencymaps}
Adebayo, J.; Gilmer, J.; Muelly, M.; Goodfellow, I.; Hardt, M.; and Kim, B.
\newblock 2018.
\newblock Sanity checks for saliency maps.
\newblock In {\em The 32Nd NeurIPS}.

\bibitem[\protect\citeauthoryear{Breuker \bgroup et al\mbox.\egroup
  }{2016}]{breuker2016comprehensible}
Breuker, D.; Matzner, M.; Delfmann, P.; and Becker, J.
\newblock 2016.
\newblock Comprehensible predictive models for business processes.
\newblock {\em MIS Quarterly}.

\bibitem[\protect\citeauthoryear{Daugherty and
  Wilson}{2018}]{daugherty2018human}
Daugherty, P.~R., and Wilson, H.~J.
\newblock 2018.
\newblock {\em Human+ machine: reimagining work in the age of AI}.
\newblock Harvard Business Press.

\bibitem[\protect\citeauthoryear{Evermann, Rehse, and
  Fettke}{2017}]{evermann2017predicting}
Evermann, J.; Rehse, J.-R.; and Fettke, P.
\newblock 2017.
\newblock Predicting process behaviour using deep learning.
\newblock {\em Decision Support Systems}.

\bibitem[\protect\citeauthoryear{Gan \bgroup et al\mbox.\egroup
  }{2015}]{gancvpr2018}
Gan, C.; Wang, N.; Yang, Y.; Yeung, D.-Y.; and Hauptmann., A.~G.
\newblock 2015.
\newblock Devnet: A deep event network for multimedia event detection and
  evidence recounting.
\newblock In {\em The 28th CVPR}.

\bibitem[\protect\citeauthoryear{Guo \bgroup et al\mbox.\egroup
  }{2018}]{wenboccs2018}
Guo, W.; Mu, D.; Xu, J.; Su, P.; Wang, G.; and Xing, X.
\newblock 2018.
\newblock Lemna: Explaining deep learning based security applications.
\newblock In {\em The 25th ACM CCS}.

\bibitem[\protect\citeauthoryear{Li \bgroup et al\mbox.\egroup
  }{2015}]{nlpinterpretability}
Li, J.; Chen, X.; Hovy, E.; and Jurafsky, D.
\newblock 2015.
\newblock Visualizing and understanding neural models in nlp.
\newblock {\em arXiv preprint arXiv:1506.01066}.

\bibitem[\protect\citeauthoryear{Mannhardt \bgroup et al\mbox.\egroup
  }{2016}]{mannhardt2016decision}
Mannhardt, F.; De~Leoni, M.; Reijers, H.~A.; and Van Der~Aalst, W.~M.
\newblock 2016.
\newblock Decision mining revisited-discovering overlapping rules.
\newblock In {\em CAiSE}.

\bibitem[\protect\citeauthoryear{Marketwatch}{2019}]{marketwatch}
Marketwatch.
\newblock 2019.
\newblock {Business Process Management (BPM) Market 2019: Key Findings,
  Regional Study, Size, Growth and Global Trends by Forecast to 2023}.
\newblock Online; accessed 2019.

\bibitem[\protect\citeauthoryear{Nguyen \bgroup et al\mbox.\egroup
  }{2016}]{nguyen2016process}
Nguyen, P.; Slominski, A.; Muthusamy, V.; Ishakian, V.; and Nahrstedt, K.
\newblock 2016.
\newblock Process trace clustering: A heterogeneous information network
  approach.
\newblock In {\em In SIAM SDM}.

\bibitem[\protect\citeauthoryear{Nguyen \bgroup et al\mbox.\egroup
  }{2019}]{nguyen2019summarized}
Nguyen, P.; Ishakian, V.; Muthusamy, V.; and Slominski, A.
\newblock 2019.
\newblock Summarized: Efficient framework for analyzing multidimensional
  process traces under edit-distance constraint.
\newblock {\em arXiv preprint arXiv:1905.00983}.

\bibitem[\protect\citeauthoryear{{Polato} \bgroup et al\mbox.\egroup
  }{2014}]{polato2014}
{Polato}, M.; {Sperduti}, A.; {Burattin}, A.; and {de Leoni}, M.
\newblock 2014.
\newblock Data-aware remaining time prediction of business process instances.
\newblock In {\em IJCNN}.

\bibitem[\protect\citeauthoryear{Rao and Verweij}{2017}]{rao2017sizing}
Rao, D. A.~S., and Verweij, G.
\newblock 2017.
\newblock Sizing the prize: What’s the real value of {AI} for your business
  and how can you capitalise?
\newblock {\em PwC Publication, PwC}.

\bibitem[\protect\citeauthoryear{Ribeiro, Singh, and
  Guestrin}{2016}]{ribeiro2016should}
Ribeiro, M.~T.; Singh, S.; and Guestrin, C.
\newblock 2016.
\newblock Why should {I} trust you?: Explaining the predictions of any
  classifier.
\newblock In {\em the 22nd ACM SIGKDD}.

\bibitem[\protect\citeauthoryear{Schielzeth}{2010}]{regressioninterpretability}
Schielzeth, H.
\newblock 2010.
\newblock Simple means to improve the interpretability of regression
  coefficients.
\newblock {\em Methods in Ecology and Evolution} 1(2):103--113.

\bibitem[\protect\citeauthoryear{Tax \bgroup et al\mbox.\egroup
  }{2017}]{tax2017predictive}
Tax, N.; Verenich, I.; La~Rosa, M.; and Dumas, M.
\newblock 2017.
\newblock Predictive business process monitoring with lstm neural networks.
\newblock In {\em CAiSE},  477--492.
\newblock Springer.

\bibitem[\protect\citeauthoryear{Van Der~Aalst and
  others}{2011}]{van2011process}
Van Der~Aalst, W., et~al.
\newblock 2011.
\newblock Process mining manifesto.
\newblock In {\em BPM}.
\newblock Springer.

\bibitem[\protect\citeauthoryear{Wilson, Alter, and
  Shukla}{2016}]{wilson2016companies}
Wilson, H.; Alter, A.; and Shukla, P.
\newblock 2016.
\newblock Companies are reimagining business processes with algorithms.
\newblock {\em Harvard Business Review}.

\end{thebibliography}
}
\end{document}